\documentclass[letterpaper, 10 pt, conference]{ieeeconf}
\IEEEoverridecommandlockouts                              % This command is only needed if 
\usepackage{graphics} % for pdf, bitmapped graphics files
\usepackage{epsfig} % for postscript graphics files
\usepackage{epstopdf}
\usepackage{amsmath} % assumes amsmath package installed
\usepackage{bm,upgreek}
\usepackage{amssymb}  % assumes amsmath package installed
\usepackage{cleveref} % refer to a range of labels
\usepackage{xcolor}
\newcommand{\ssin}[1]{{s}_{(#1)}}   % shorthand sin
\newcommand{\scos}[1]{{c}_{(#1)}}   % shorthand cos

\usepackage[stable]{footmisc}  % for footnote
\usepackage{cite}
\title{\LARGE \bf Towards Cooperative Transport of a Suspended Payload \\via Two Aerial Robots with Inertial Sensing
}

\author{Heng Xie, Xinyu Cai, and Pakpong Chirarattananon
\thanks{This work was supported by the Research Grants Council of the Hong Kong Special Administrative Region of China (grant number CityU-11215117).}% <-this % stops a space
\thanks{The authors are with the Department of Biomedical Engineering, City University of Hong Kong, Hong Kong SAR, China (email: pakpong.c@cityu.edu.hk).}
}%

\begin{document}
\maketitle

\begin{abstract}
This paper addresses the problem of cooperative transport of a point mass hoisted by two aerial robots. Treating the robots as a leader and a follower, the follower stabilizes the system with respect to the leader using only feedback from its Inertial Measurement Units (IMU). This is accomplished by neglecting the acceleration of the leader, analyzing the system through the generalized coordinates or the cables' angles, and employing an observation model based on the IMU measurements. A lightweight estimator based on an Extended Kalman Filter (EKF) and a controller are derived to stabilize the robot-payload-robot system. The proposed methods are verified with extensive flight experiments, first with a single robot and then with two robots. The results show that the follower is capable of realizing the desired quasi-static trajectory using only its IMU measurements. The outcomes demonstrate promising progress towards the goal of autonomous cooperative transport of a suspended payload via small flying robots with minimal sensing and computational requirements. 
\end{abstract}

\section{Introduction}

In the past decades, we have witnessed a rapid development of small flying robots \cite{chen2019controlled,hsiao2019ceiling}. The development of miniaturized electronics and associated algorithms allows these Micro Aerial Vehicles (MAVs) to be used for surveillance, mapping, agriculture, delivery, etc. Owing to their small footprint, more recently, there has been a growing interest in the use of these small drones in collaborative tasks, including for swarm behavior~\cite{mulgaonkar2017robust}, manipulation~\cite{estrada2018forceful} or collective transportation of heavy objects either by rigid attachments~\cite{mu2019universal,loianno2017cooperative,morin2019autonomous,tagliabue2019robust} or suspension~\cite{sreenath2013dynamics,gassner2017dynamic}.

This paper focuses on the cooperative transport of a suspended payload by multiple MAVs. In early stages, the problem of aerial transport of a suspended payload has been considered in the single-robot context~\cite{murray1996trajectory,sreenath2013trajectory,tang2015mixed,yu2019nonlinear, de2019two}. To increase the load-carrying capacity, uses of multiple vehicles emerge as an appealing solution~\cite{sreenath2013dynamics,loianno2017cooperative}. However, the strategy inevitably results in additional complications as the dynamics of multiple robots and the payload become coupled.

As a result, studies concerning cooperative manipulation of a payload suspended from multiple robots focus on the dynamics, trajectory generation, and control~\cite{sreenath2013dynamics,michael2011cooperative,gassner2017dynamic}. In~\cite{sreenath2013dynamics}, the authors showed that the system of multiple robots carrying a point mass or rigid body payload is differentially flat when the trajectory of the payload is chosen as the flat output. The framework assists in the computation of the trajectories of associated robots, allowing the collective transport task to be realized in actual experiments. Nevertheless, similar to other related works~\cite{michael2011cooperative, lee2017geometric, shirani2019cooperative}, the proposed solution necessitates comprehensive feedback of robots' pose and global communication, limiting the application to  laboratory environments with a motion capture system and global communication as found in~\cite{michael2011cooperative} or simulation as shown by~\cite{lee2017geometric,shirani2019cooperative}. 

The feedback and communication issue has recently been tackled by incorporating vision~\cite{gassner2017dynamic} (and~\cite{loianno2017cooperative,tagliabue2019robust} for a rigidly attached payload). The method in~\cite{gassner2017dynamic} employs a leader-follower approach. This requires two robots to visually track the transported object and the follower to directly observe the leader. Simiarly, vision has been used to estimate the pose of the robot-payload-robot formation \cite{loianno2017cooperative} or to estimate the force applied to the payload by the other robot ~\cite{tagliabue2019robust}. While these approaches do not require explicit communication between agents, the deployments rely heavily on visual feedback, amplifying the computational cost and sensory payload, rendering it unsuitable for smaller flying robots.

 \begin{figure}[t]
     \centering
     { \includegraphics[width=.48\textwidth]{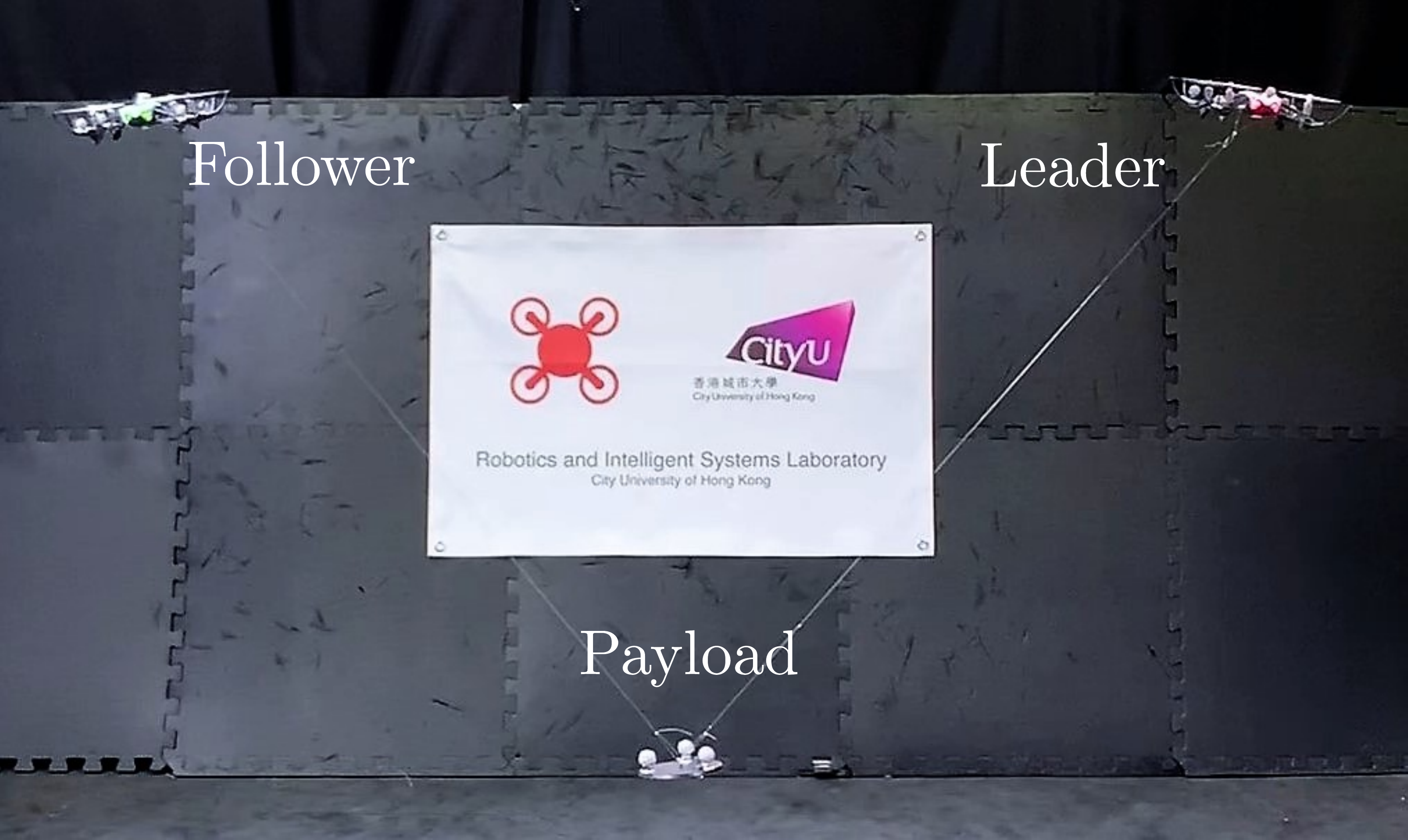}}
     \caption{Two robots in a leader-follower configuration transporting a point-mass payload. The formation is stabilized by the follower robot using only feedback from its IMU.}
     \label{fig:cooperHoverPhoto}
 \end{figure}
This work offers an alternative strategy for the transportation of a point-mass payload hoisted by two robots as portrayed in Fig.~\ref{fig:cooperHoverPhoto}. The proposed method requires only the IMU and no communication between two robots. The limitation to two-robot scenarios let us treat the robots as a leader and a follower. By restricting to non-aggressive maneuvers or neglecting the acceleration of the leader (this does not prevent the robots to traverse at moderate or high speed), the desired quasi-static trajectory is essentially stabilized by the follower. To achieve this, the follower robot uses only IMU feedback to estimate the state, or its relative position to the leader, via the developed EKF-based estimator. Then, a controller is derived based on the linearized dynamics. To this end, the state dynamics are stabilized according to the estimated state vector computed entirely from the IMU measurements. Despite some constraints in the current implementation, this framework provides a novel approach for aerial collective transport with minimal sensing requirements.

%In this work, we address the leader-follower formation control problem of cooperative transporting a payload that tethered by two flying vehicles, as shown in Fig~\ref{fig:cooperHoverPhoto}. We assume the leader keeps hover and strings are always taut, then explore the relationship between the system's formation and the follower's position and attitude, we learn that with the knowledge of thrust, the follower is able to know its relative position and velocity to the leader and the payload by only relying on its IMU, without any direct communication to the leader. Hence we access the thrust through a first-order state-space model built by using subspace identification, and then design a nonlinear Kalman Filter to estimate the formation's state successfully by using the inertial sensing information. The estimation for the formation angles extends our perception for what an IMU system can do on a robot, and it not only provides attitude information for the robot but also indicates knowledge about the outer system beyond itself under certain circumstances. Also, we develop a linear controller to drive the follower such that the system can track a predefined formation by using the estimated state as feedback. 

This paper is organized as follows. In Section~\ref{sec:systemModeling}, the simplified dynamics of leader-payload-follower system are given. This includes the observation model that leads to the derivation of the state estimator and controller in Section~\ref{sec:estimation}. Flight experiments are perfomed in Section~\ref{sec:results} to evaulate the performance of the proposed estimator and controller. Conclusion and discussion of possible extensions of this work are given in Section~\ref{sec:conclusion}.

\section{Leader-Follower System Dynamics} \label{sec:systemModeling}
This work consider a transportation of single payload suspended between two aerial robots in the leader-follower manner with cables are presumed massless and always taut. To simplify the consideration, the leader's dynamics are assumed quasi-static---its acceleration is negligible. Additional aerodynamic damping forces are also neglected. Presently, this limits the operating conditions to low-to-moderate speed flights with small accelerations.

\subsection{Reduced Dynamic Model} \label{subsec:dynamicModel}
\begin{figure}[t]
    \centering
    \includegraphics[width=.45\textwidth]{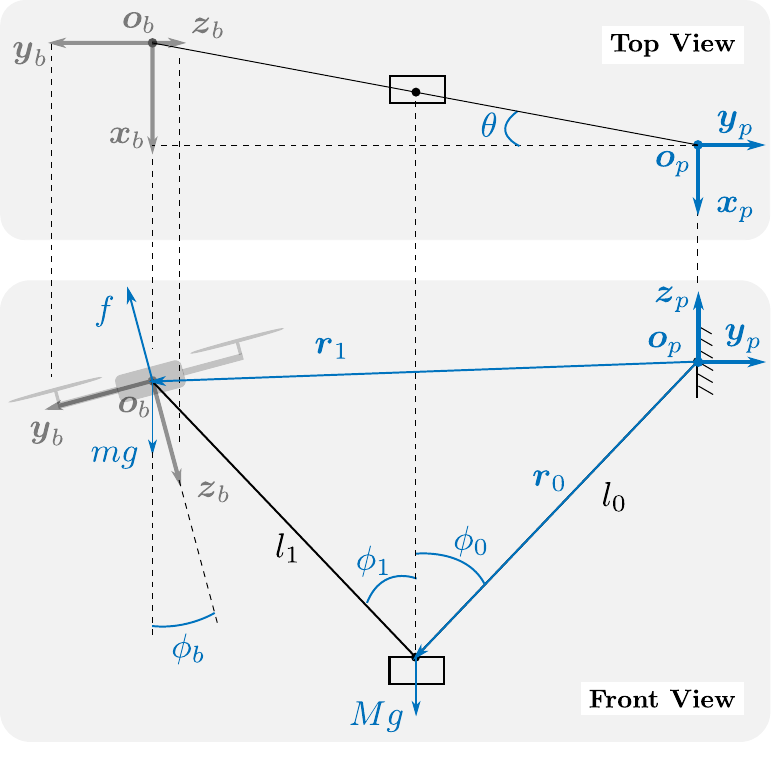}
    \caption{(Front view) A simplified system in 2D. The payload is tethered to a follower and a fixed-point (representing a leader). (Top view) The robot and payload may deviate from the $y_p$-$z_p$ plane, in which case the component of cable tension along $x_p$ axis would provide a restoring force pulling the robot back towards the $y_p$-$z_p$ plane.}
    \label{fig:2-DsystemModel}
\end{figure}

Fig.~\ref{fig:2-DsystemModel} illustrates planar views of the coordinate frames and associated parameters of the two-robot system. The origin of frame $\mathcal{P}:\{x_p,y_p,z_p\}$ is assumed coincident with the center-of-mass (CoM) of the leader. Thanks to the near-hovering assumption, frame $\mathcal{P}$ is regarded as an inertial frame, allowing the dynamics of the leader to be neglected. Frame $\mathcal{B}:\{x_b, y_b,z_b\}$ is a body-fixed frame associated with the follower. The follower, with mass $m$, generates a collective thrust $f$ in the direction opposite to $z_b$. The payload, with mass $M$, is a point mass hoisted between two robots by massless cables with lengths $l_0$ and $l_1$. The cable's attachment point passes through the CoM of the robot. Consequently the cable tension makes no contribution to the attitude dynamics of the follower. Positions of the payload and the follower with respect to $\mathcal{P}$, denoted by $\bm{r}_0$ and $\bm{r}_1$, are described by the angles of the cable $\phi_0$ and $\phi_1$ measured from the vertical about $x_p$. Since the pitch and yaw dynamics of the follower are not directly affected by the suspended payload, at this stage, they are assumed to be separately controlled such that the yaw angle is always zero (hence, $y_b$ is parallel to $y_p$). The out-of-plane motion (indicated by $\theta$ in Fig.~\ref{fig:2-DsystemModel}) is separately minimized by the robot's pitch control so that $\theta\approx 0$ as outlined in Section~\ref{subsec:deviation} below.

We let $\phi_b$ denote the roll angle (or rotation about $y_b$) of the robot, the attitude and translational dynamics of the robot and the payload are fully described by the generalized coordinates $\Phi = [{\phi}_b, {\phi}_0, {\phi}_1 ]^{T}$ and can be derived from the Euler-Lagrange equation and corresponding the Lagrangian:
\begin{align}
 \bm{\tau} =& \frac{\text{d}}{\text{d}t}\left( \frac{\partial \mathcal{L}}{\partial \dot{\Phi}}  \right) -\frac{\partial \mathcal{L}}{\partial \Phi} \label{equ:euler-lagrange} \\
\mathcal{L}(\Phi,\dot{\Phi}) =& \mathcal{K}(\Phi,\dot{\Phi})-\mathcal{U}(\Phi) \label{equ:lagrangian}
\end{align}
% In this work, the leader is supposed to be always hovering at a fixed point such that we can model the leader as an anchor. Aerodynamics and masses of strings are neglected and assume they are always taut. Define a vertical plane $\{o_p, y_p, z_p\}$, which is a plane of frame $\mathcal{P}:\{x_p,y_p,z_p\}$, assuming axis $x_b$ of the body frame $\mathcal{B}$, $\{x_b, y_b,z_b\}$ is always coincided with $x_p$ of $\mathcal{P}$, as shown in the Front View of Fig.~\ref{fig:2-DsystemModel}, we simplify the problem into a 2D planar case: A payload is tethered by two strings, one is connected to a flying quadrotor and the other one is attached to a fixed point, which results to a 'V' like formation (note that the word 'formation' throughout of this work are referred to this 'V' in the vertical 2D plane). We develop dynamic and measurement models of the system, to design estimation and control algorithms.
where
\begin{align}
\mathcal{K}&= \frac{1}{2}M\dot{\bm{r}}_0^T\dot{\bm{r}}_0 + \frac{1}{2}m\dot{\bm{r}}_1^{T}\dot{\bm{r}}_1 + \frac{1}{2}\bm{I}_b \dot{\phi}_b^2 \\
\mathcal{U} &= \begin{bmatrix}
0 & 1
\end{bmatrix} \left ( Mg \bm{r}_{0} + mg \bm{r}_{1} \right )  \\
\bm{r}_0 &=\begin{bmatrix}
l_0 \ssin{{\phi}_0} \\ - l_0\scos{{\phi}_0}
\end{bmatrix}, \ 
%\dot{\bm{r}}_0 &= \begin{bmatrix}
%\dot{\phi}_0 l_0 \cos{{\phi}_0} \\ \dot{\phi}_0 l_0 \sin{{\phi}_0}
%\end{bmatrix} 
\bm{r}_1 = \begin{bmatrix}
l_0\ssin{{\phi}_0} - l_1\ssin{{\phi}_1} \\ - l_0\scos{{\phi}_0} + l_1\scos{{\phi}_1} \label{equ:euler_lagrange_r1}
\end{bmatrix}
%\dot{\bm{r}}_1 &= \begin{bmatrix}
%\dot{\phi}_0 l_0 \cos{{\phi}_0} - \dot{\phi}_1 l_1 \cos{{\phi}_1} \\
%\dot{\phi}_0 l_0 \sin{{\phi}_0} - \dot{\phi}_1 l_1 \sin{{\phi}_1}
%\end{bmatrix} 
\end{align}
c$_{(\cdot)}$ and s$_{(\cdot)}$ are shorthands for $\cos{(\cdot)}$ and $\sin{(\cdot)}$, $\bm{I}_b$ is the moment of inertia about the roll axis the follower, and $\bm{\tau}$ is the generalized torque evaluated according to D'Alembert's principle. Solving Eq.~\eqref{equ:euler-lagrange}-\eqref{equ:euler_lagrange_r1} yields the manipulator equation
\begin{equation}
{H}\left ( \Phi \right )\ddot{\Phi} + {C}\left ( \Phi,\dot{\Phi} \right ) + {G}\left ( \Phi \right ) = {B}\left ( \Phi \right )\bm{u} \label{equ:manipulator}
\end{equation}
with the following definitions:
\begin{align}
{H} &= \begin{bmatrix}
1 & 0 & 0 \\ 
0 & (m+M)l_0^2 & -ml_0l_1\scos{\phi_0-\phi_1} \\ 
0 &  -ml_0l_1\scos{\phi_0-\phi_1} & ml_1^2
\end{bmatrix} \nonumber \\
{C} &= \begin{bmatrix}
0\\ 
-ml_0l_1\dot{\phi}_1^2\ssin{\phi_0-\phi_1}\\ 
ml_0l_1\dot{\phi}_0^2\ssin{\phi_0-\phi_1}
\end{bmatrix}, \
{G} =  \begin{bmatrix}
0\\ 
(m+M) gl_0\ssin{\phi_0}\\
-mgl_1\ssin{\phi_1}
\end{bmatrix} \nonumber \\
{B} &= \begin{bmatrix}
1 & 0\\ 
0 & l_0 \ssin{\phi_0-\phi_b} \\ 
0 & l_1 \ssin{\phi_b-\phi_1}
\end{bmatrix}, \text{where }\ 
\bm{u} = \begin{bmatrix}
\tau_b\\ 
f
\end{bmatrix} \label{equ:manipulator_defs}
\end{align}
is the system's input. Finally, the nonlinear dynamics described by Eq.~\eqref{equ:manipulator}-\eqref{equ:manipulator_defs} can also be expressed using the state vector $\bm{x} = [{\phi}_b, \dot{\phi}_b, {\phi}_0, \dot{\phi}_0, {\phi}_1, \dot{\phi}_1]^{T}$ as a first-order differential equation:
\begin{equation} 
    \dot{\bm{x}} = \bm{f}(\bm{x},\bm{u}) \label{equ:state_dynamics}
\end{equation}
\subsection{Measurement Model}
Since no external feedback or additional sensors are employed, the measurements available for the follower are strictly from the onboard IMU. Since the state dynamics defined by Eq.~\eqref{equ:state_dynamics} are planar, we define the output vector
\begin{equation}
\bm{y} = \begin{bmatrix}
{\phi}_b \quad \dot{\phi}_b \quad a_{y} \quad a_{z}
\end{bmatrix}^{T} \label{equ:output_vector}
\end{equation}
where the angle $\phi_b$ and the angular rate $\dot{\phi}_b$ can be provided the complementary filter or other low level architecture, and $a_{y}$ and $a_{z}$ are the accelerometer readings along the $y_b$ and $z_b$ axes of the robot.

To obtain the measurement model, $\bm{y}=\bm{h}(\bm{x},\bm{u})$, we first consider a minimal model of the complementary filter. Neglecting the inertial term and high-frequency dynamics, the measurement value of  roll angle is approximated as the ratio of $a_{y}$ to the standard gravity or $\phi_b\approx -\sin^{-1}(a_{y}/g)$. The gravity-subtracted acceleration, $a_{y}$ and $a_{z}$, are obtained from the vector summation of the collective thrust ($f$) and the tension in cable $l_1$ ($T_1$), normalized by the robot's mass ($m$) according to $ma_{y}=-T_1\ssin{{\phi}_1-{\phi}_b}$ and $ma_{z}=T_1\scos{{\phi}_1-{\phi}_b}-f$. Meanwhile, $T_1$ can be computed by analysing the torque dynamics of $\phi_0$ about $x_p$, which yields 
\begin{align} 
Ml_0^2\ddot{\phi}_0 = T_1l_0 \ssin{{\phi}_0 - {\phi}_1}-  Mgl_0\ssin{{\phi}_0} \label{equ:cable_tension}
\end{align} 
Subsequently, the measurement model is given as
\begin{equation} \label{equ:outputModel}
\bm{y} = \bm{h}(\bm{x},\bm{u})
 = \begin{bmatrix}
\phi_b \\ 
    \dot{\phi}_b \\ 
    -(T_1/m)\ssin{{\phi}_1-{\phi}_b} \\ 
    (T_1/m)\scos{{\phi}_1-{\phi}_b}-(f/m)
 \end{bmatrix}
\end{equation}
where $T_1$ is evaluated from Eq.~\eqref{equ:cable_tension}, and $\ddot{\phi}_0$ from the state dynamics Eq.~\eqref{equ:state_dynamics}. As a result, $\bm{h}(\bm{x},\bm{u})$ relates the state and input vectors to the measurement $\bm{y}$ as required.

Notice that without ignoring the acceleration of the leader, it would contribute to the system's dynamics as an additional system's input. Without the knowledge of such term (via direct measurements or communication between the leader and the follower), the state would be unobservable. This illustrates the importance of the employed quasi-static assumption.

\subsection{Out-of-plane Dynamics} \label{subsec:deviation}
Thus far, the dynamic models of the state and measurement have been derived by neglecting the out-of-plane motion or assuming the angle $\theta$ (defined about $z_p$) in Fig.~\ref{fig:2-DsystemModel} is small and the yaw angle is controlled. As a result, the position of the follower, as determined by $\phi_0$ and $\phi_1$, is only dependent on the robot's roll dynamics, uncoupled from the out-of-plane motion. Here, we inspect this out-of-plane dynamics under the condition $\left |\theta \right |\ll 1$ to verify that this assumption can be satisfied in practice.

To begin, we recall that the onboard controller actively minimizes the follower's pitch angle. This means that the collective thrust lies approximately in the $\{o_p,y_p,z_p\}$ plane. As seen in Fig.~\ref{fig:2-DsystemModel}, the motion of the robot along $x_p$ is governed by the cable tension $T_1$, that is
\begin{equation}
\begin{aligned}
    m\ddot{x}_b & = -T_1\ssin{\phi_1}\tan{\theta} \label{equ:out-of-plane-dyn}
\end{aligned}
\end{equation}
Near the equilibrium state, we assume $\phi_0$, $\phi_1$ and $T_1$ are constant. With the small angle approximation,  $\tan\theta\approx \theta$ and $\ddot{x}_b\approx (-l_0\ssin{{\phi}_0} + l_1\ssin{{\phi}_1}) \ddot{\theta}$. Eq.~\eqref{equ:out-of-plane-dyn} becomes
\begin{equation}
  \ddot{\theta} = -\frac{T_1\ssin{\phi_1}}{m(-l_0\ssin{{\phi}_0} + l_1\ssin{{\phi}_1})} \theta
\end{equation}
In a normal operating range, it is anticipated that $\phi_0 \in (-\frac{\pi}{2},0)$ and $\phi_1 \in (0,\frac{\pi}{2})$. The tension $T_1$ then provides a restoring torque, resulting in a marginally-stable simple harmonic oscillation. In practice, unmodeled aerodynamic damping likely produces the desirable stabilizing effect. In other words, controlling the robot's pitch and yaw angles is sufficient to ensure $\theta\rightarrow 0$, minimizing the out-of-plane motion.
  
\begin{figure*}[tbp]
    \centering
    \includegraphics[width=.84\textwidth]{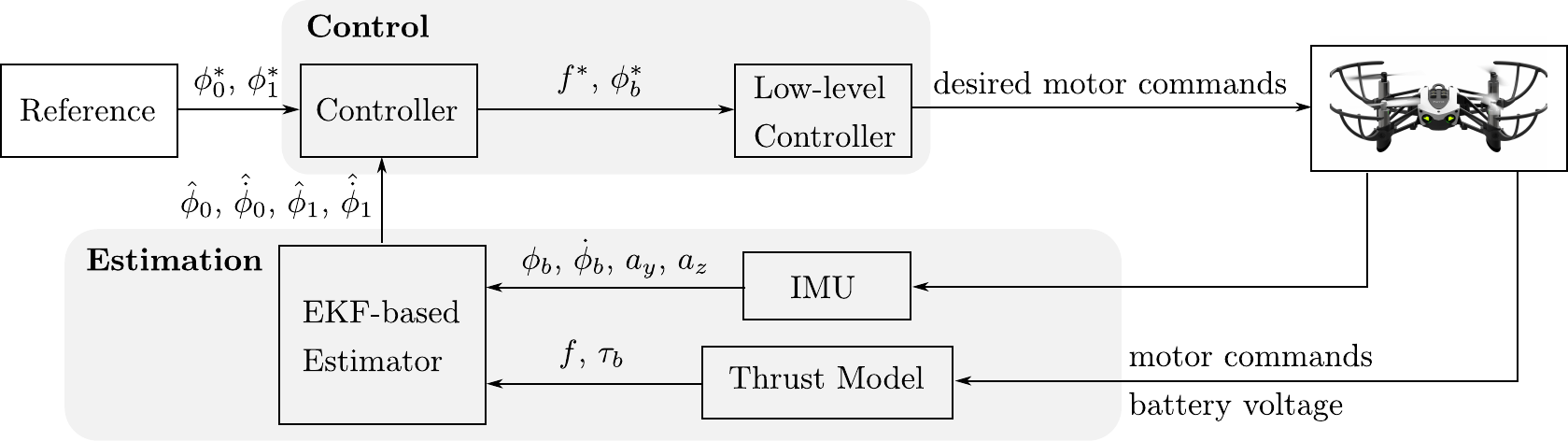}
    \caption{The onboard estimation and control architecture. The state feedback is provided by an EKF-based estimator. The state controller computes the desired collective thrust and roll angle that can be directly used for generating motor commands by the low-level controller of the follower quadrotor.}
    \label{fig:formationControlArchi}
\end{figure*}

\section{State Estimation \& Control} \label{sec:estimation}
With the state dynamics and measurement model, in this section, we describe the estimation and control strategy to control the position of the follower robot, or realizing the $\phi_0$ and $\phi_1$ setpoints. To achieve this, an EKF-based estimator is proposed, allowing the state vector to be estimated from the IMU measurements. Then, a controller is derived based on a decoupled linearized state dynamics. The architecture of the estimator, controller, as well its their connections to the low-level flight controller is depicted in Fig.~\ref{fig:formationControlArchi}.
 %We design Kalman Filters to estimate the state $\bm{x}$, we have tried \textit{Extended Kalman Filter} (EKF) and \textit{Unscented Kalman Filter} (UKF). The EKF linearizes all nonlinear equations about current estimated state and substitutes Jacobian matrices for the linear transformations in the KF equations, while the UKF propagates deterministically chosen sigma points that represent the mean and covariance information through nonlinear transformations, the Kalman gain computation, and state correction are the same as in KF~~\cite{simon2006optimal}. It turns out that in this case, UKF estimation accuracy is better than EKF, but it is less efficient~~\cite{julier2004unscented}. Because of the limited computational power of the hardware platform and high requirement of estimation accuracy for control purpose, we finally design a \textit{Single Propagation Unscented Kalman Filter} (SPUKF), which only propagates the mean through nonlinear transform, then the sigma points are approximated by first-order Taylor Series about the transformed mean point, results show that it is comparable to standard UKF in terms of estimation accuracy, also is comparable to EKF in terms of computational efficiency~~\cite{biswas2016novel}.
\subsection{Observability and Controllability} \label{subsec:obsv-ctrl}
To ensure that the state vector can be observed and controlled according to the definitions defined by Eq.~\eqref{equ:state_dynamics} and \eqref{equ:output_vector}, the dynamic model is linearized about a nominal state and verified for observability and controllability. The results show that the rank conditions of both observability and controllability matrices are satisfied.

\subsection{State Estimation via EKF-based estimator}
To estimate the state vector from available measurements, a filter or an estimator must be employed. A common solution for nonlinear dynamics and observation models is an Extended Kalman Filter, thanks to its simple implementation. However, our preliminary study reveals that EKF is unable to provide estimates with satisfactory accuracy and reliability. This is possibly due to the linearization that results in inaccurate covariance propagation. This shortcomings can be addressed by an Unscented Kalman Filter, which provides better estimation of the posterior distribution by deterministic sampling, improving the robustness and accuracy over the EKF ~\cite{gustafsson2011some}. Nevertheless, computational cost of the UKF is an order of magnitude higher than EKF~\cite{simon2008comparison}, making the UKF unsuitable for real-time onboard implementation on a small flying robot platform without a companion computer. To this end, we incorporated an extra tuning parameter $\lambda$ into the standard EKF. Motivated by UKF, this $\lambda$ plays an identical role to the parameter that controls the spread of the sigma points in UKF ~\cite{gustafsson2011some} and can be seen as a scaling parameter for adjusting the error covariance during the state prediction and update. Choosing $\lambda$ to be less than unity, for instance, improves the estimator's performance when the linearization overesimates the covariance. Compared to UKF, the proposed EKF-based estimator offers rivaled performance to UKF with a similar computational requirement to EKF.

Here, we briefly describe the implementation of an EKF-based estimator. This begins by expressing the state and observation functions (Eq.\eqref{equ:state_dynamics} and \eqref{equ:outputModel}) in the discrete-time domain using the forward Euler method.
\begin{equation}
\begin{aligned}
\bm{x}_k &= \bm{x}_{k-1} + \bm{f}(\bm{x}_{k-1},\bm{u}_k)\Delta T + \bm{w}_k \\
\bm{y}_k &= \bm{h}(\bm{x}_k,\bm{u}_k) + \bm{v}_k 
\end{aligned}
\end{equation}
where $\Delta T$ is a sample time, $k$ denotes the time index at instant $t_{k}$, $\bm{w}_k$ and $\bm{v}_k$ are zero-mean Gaussian process and measurement white noises with the associated covariance matrices $\bm{Q}_k \in \mathbb{R}^{6 \times 6}$ and $\bm{R}_k \in \mathbb{R}^{4 \times 4}$. The prediction and update procedures follow closely those of standard EKF as follows.
\subsubsection{Prediction}
The prediction step resembles a standard form
\begin{align}
	\hat{\bm{x}}_k^- &= \hat{\bm{x}}_{k-1}^+ + \bm{f}(\hat{\bm{x}}_{k-1}^+,\bm{u}_k) \Delta T \\
	\bm{\Sigma}_{\bm{x},k}^{-} &= \lambda\bm{F}_k\bm{\Sigma}_{\bm{x},k-1}^{+}\bm{F}_k^T + \bm{Q}_k
\end{align} 
where $\hat{\bm{x}}_{k-1}^+$ is an \textit{a-posteriori} estimate at time $t_{k-1}$, $\hat{\bm{x}}_k^-$ an \textit{a-priori} estimate at time $t_k$, $\Sigma_{\bm{x}}$ is a state covariance estimate, $\bm{F}_k$ is the state Jacobian $\partial \bm{f}/\partial\bm{x}|_{\hat{\bm{x}}^+_{k-1},\bm{u}_k}$, and $\lambda$ is an extra scalar tuning parameter.
\subsubsection{Update}
The innovation covariance $\bm{\Sigma}_{\bm{y},k}$ and Kalman gain $\bm{K}_k$ are also modified to include $\lambda$ as
\begin{align}
	\bm{\Sigma}_{\bm{y},k} &= \lambda\bm{H}_k\bm{\Sigma}_{\bm{x},k}^{-}\bm{H}_k^T + \bm{R}_k \\
	\bm{K}_k &= \lambda \bm{\Sigma}_{\bm{x},k}^{-}\bm{H}_k^T \bm{\Sigma}_{\bm{y},k}^{-1}
\end{align} 
where $\bm{H}_k$ is the observation Jacobian $\partial \bm{h}/\partial\bm{x}|_{\hat{\bm{x}}^-_{k},\bm{u}_k}$. Subsequently, the updated state and covariance estimates are
\begin{align}
	\hat{\bm{x}}_{k}^{+} &= \hat{\bm{x}}_{k}^{-} + \bm{K}_k\left(\bm{y}_k - \bm{h}(\hat{\bm{x}}_k^-,\bm{u}_k) \right) \\
	\bm{\Sigma}_{\bm{x},k}^{+} &= \bm{\Sigma}_{\bm{x},k}^{-} - \lambda \bm{K}_k\bm{H}_k \bm{\Sigma}_{\bm{x},k}^{-}.
\end{align}

\subsection{State Control} \label{sec:control}

To realize cooperative transportation in the leader-follower manner by tethered flying robots, it is vital to control the cables' angles. In this part, several simplifying assumptions are employed. First, the state vector is assumed known from the estimation scheme described above. Next, the roll dynamics of the robot are presumed considerably faster than the dynamics of the tethered system. This approximation decouples the dynamics of $\phi_b$ from $\phi_0$ and $\phi_1$, allowing the robot's thrust $f$ and its roll angle $\phi_b$ to be treated as inputs of the simplified system ($\bm{u}_r = [f, \phi_b]^{T}$). The system's dynamics previously given by Eq.~\eqref{equ:manipulator} reduce to
\begin{equation}
{H}_r\left ( \Phi \right )\ddot{\Phi} + {C}_r\left ( \Phi,\dot{\Phi} \right ) + {G}_r\left ( \Phi \right ) = U_r \label{equ:reduced_manipulator}
\end{equation}
where ${H}_r$ is constructed from the $2\times2$ bottom right elements of $H$, ${C}_r$ and ${G}_r$ are the taken as the last two rows of $C$ and $G$, and $U_r=[f l_0 \ssin{\phi_0-\phi_b},f  l_1 \ssin{\phi_b-\phi_1} ]^T$.

For the desired constant setpoints: $\phi_0^*$, $\phi_1^*$, the corresponding feedforward input $\bm{u}_r^* = [f^*, \phi_b^*]^{T}$ can be found from solving the equation $G_r|_{\phi_0^*,\phi_1^*} = U_r|_{\phi_0^*,\phi_1^*,f^*,\phi_b^*}$.

To stabilize the system, Eq.~\eqref{equ:reduced_manipulator} is linearized about the nominal conditions: $\phi_{0,1}=\phi_{0,1}^*+\Delta\phi_{0,1}$, $\bm{u}_r=\bm{u}_r^*+\Delta\bm{u}_r$, and $\dot{\phi}_{0,1}=0$, this yields 
\begin{equation}
{H}_r^*\Delta \ddot{\Phi} + {P_r}^{*} \Delta\Phi  = {B}_r^*\Delta \bm{u}_r \label{equ:linearizedSystem}
\end{equation}
where
\begin{equation}
P_r^* = \frac{\partial}{\partial \Phi}\left (G_r-U_r \right)  |_{\Phi^*,u_r^*}, \quad B_r^* = \frac{\partial}{\partial \bm{u}_r} U_r |_{\Phi^*,u_r^*}
\end{equation}
It turns out that both $H_r^*$ and $P_r^*$ are positive definite for $-\phi_0^*,\phi_1^*\in (0, \frac{\pi}{2})$, or the linearized system is marginally stable. 
The structure and stability property of Eq.~\eqref{equ:reduced_manipulator} suggests that it can be stabilized by feedback in the standard PID form 
\begin{equation}
(H_r^{-1}{B}_r)^*\Delta \bm{u}_r = 
-K_{p}\Delta\Phi - K_{d}\Delta\dot{\Phi} - K_{i}\int\Delta\Phi\text{d}t  \label{equ:linear_control_law}
\end{equation}
using the fact that $H_r^{*}$ and $B_r^*$ are always invertible as $|H_r^{*}| = (m^2\ssin{\phi_0^*-\phi_1^*}^2 + mM)(l_0l_1)^2 \neq 0$, and $|B_r^*|  = f^*l_0l_1\ssin{\phi_0^*-\phi_1^*} \neq 0$ at all operating points.

As a result, the follower robot is directly commanded to generate the collective thrust $f=f^*+\Delta f$, whereas the desired roll setpoint $\phi_b=\phi_b^*+\Delta \phi_b$ is given to the onboard low-level attitude controller. 

It can be seen that the decision to decouple the robot's attitude dynamics from the rest of the system radically simplifies the implementation, permitting two subsystems to be separately controlled as illustrated in Fig.~\ref{fig:formationControlArchi}. The onboard implementation is achieved with the supplement of the EKF-based estimator and the control law described by Eq.~\eqref{equ:linear_control_law}, with minimal changes to the existing attitude controller.

\section{Flight Experiments and Results} \label{sec:results}
To evaluate the proposed estimation and control methods, several flight experiments were conducted. First, we performed flights with a single robot, substituting the leader robot with a fixed point as assumed in the model derivation. Three sets of experiments are given to separately verify the estimation and control strategies before combining them. In the end, dual-robot experiments, in which the leader robot is present and manually controlled to follow low-speed trajectories by a human operator, are also provided.

%the performance of estimator can be analyzed by the open-loop experiment, which means, constant inputs at equilibriums are fed into the system without feedback control. At the same time, the system's open-loop behavior we discussed in Subsection~\ref{subsec:LocalStability} can be tested. Secondly, implement the controller not using onboard estimated state feedback, but using feedback information from motion capture system, this helps to let us focus on controller's design and to be free with possible influence from estimation. Then, we combine estimator and controller onboard to the follower quadrotor, test them with a fixed point acting as the leader. And finally, substitute the fixed point with a remotely controlled hovering quadrotor.

% We repeat tree trails for each kind of experiments mentioned above, $\phi_{1}=40^{\circ}$ and $\phi_0 = -40^{\circ}$ are chosen as the reference formation angles through out all the experiments. The equilibrium control inputs, $f^{*}$ and $\phi^{*}_b$ can be computed according to desired equilibrium angles by using equations~\eqref{equ:angleDynamics}, ~\eqref{equ:controller} and~\eqref{equ:controlInputs}.

\subsection{Experimental Setup}
Flight experiments were carried out in an indoor arena equipped with six Prime 13w motion capture (MOCAP) cameras (NaturalPoint, OptiTrack) for providing groundtruth measurements. Two Mambo Minidrones (Parrot SA) are used as the leader and follower. Each robot's mass is 73 g including markers for the MOCAP cameras. The 3D-printed dummy payload with markers weighs 30 g---a significant amount with respect to the robot's weight. It is suspended by two inelastic nylon fishing lines with 0.35 mm diameter. The cable lengths are $l_0=94$ cm and $l_1=95$ cm for all experiments. The estimation and control algorithms are implemented onboard using Simulink with the Support Package for Parrot Minidrones (MathWorks)
and executed at the nominal frequency of 200 Hz.
% When required, real-time communication with the ground station is achieved via bluetooth. Flight data are locally logged and available for offline debugging or troubleshooting after flights. 
The setpoint angles were chosen as $\phi_0=-40^\circ$ and $\phi_1=40^\circ$ throughout to strike the balance between  efficiency (favoring small angles) and robustness (large angles preferred to avoid possible collisions). Pitch and yaw angles were directly controlled by the existing low-level attitude controller without modification. Each flight contains a total of 60 s of flying time, including 5 s for taking off and landing at the beginning and the end.

\subsection{Implementation in 3D Space}

\begin{figure}[tb]
    \centering
    \includegraphics[width=.45\textwidth]{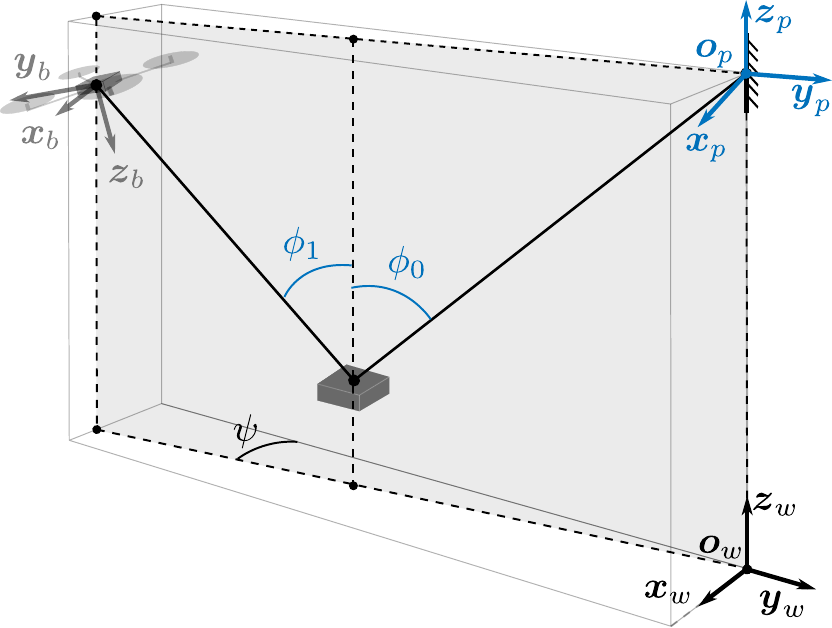}
    %\vspace*{-6mm}
    \caption{Ground truth of state in 3D experimental situation. The plane $y_p$-$z_p$, in which the system is, could rotate about $z_w$ axis due to robot's yaw motion, such that there could be an angle $\psi$ between $y_p$-$z_p$ and $y_w$-$z_w$.} \label{fig:groundTruth}
\end{figure}

The consideration of dynamics presented in Section \ref{sec:systemModeling} assumes that the robot's yaw angle is constant and zero. This permits us to consider the dynamics of the robot on a plane parallel to $y_b$ and $x_b$ axes in Fig.~\ref{fig:2-DsystemModel}. In practice, the true yaw angle is not always zero the drift of onboard feedback, resulting in a non-zero $\psi$ angle in the inertial frame $\mathcal{W}$ as shown in Fig.~\ref{fig:groundTruth}. To resolve this discrepancy, frame $\mathcal{P}$ is continuously redefined according to $\psi$, rendering $x_p$ to always be parallel to $x_b$ as previously assumed in \ref{sec:systemModeling}.

The groundtruth values of $\phi_0$ and $\phi_1$ are calculated from the projected plane perpendicular to $x_b$ and $x_p$ regardless of the actual yaw angle. Subsequently, $\dot{\phi}_0$ and $\dot{\phi}_1$ are taken as their filtered derivatives.

\subsection{Collective Thrust Model}
Since both estimator and control strategies necessitate the knowledge of thrust as one of the system inputs, this collective force is systematically identified prior to flight experiments by mounting the robot on a loadcell (Nano25, ATI). Details of the measurement procedure are similar to the process in ~\cite{hsiao2019ceiling}. The collective thrust is modeled as a function of the motor commands and battery voltage (as provided by the onboard sensor of a Minidrone) using a $0^{\text{th}}$-order model. The respective model coefficients are obtained by fitting the force sensor measurements using the \textit{n4sid} algorithm ~\cite{favoreel2000subspace}. 

\subsection{Single-Robot Experiments}
In the first three sets of experiments, the leader robot is replaced with a fixed structure and frame $\mathcal{P}$ becomes a true inertial frame.

\subsubsection{Open-loop flights}
To verify the estimation scheme, three open-loop flights were carried out. This means that the control law described in Section~\ref{sec:control} only produces the feedforward input $\bm{u}_r^*$ with no corrective terms, resulting in a constant, non-zero $\phi_b^*$. This is distinct from an unmodified control law, which would attempt to stabilize the robot to the roll setpoint $\phi_b^*=0^\circ$.

These open-loop experiments are designed not only to validate the performance of the EKF-based estimator, but also to verify that, without the corrective terms in Eq.~\eqref{equ:linear_control_law}, the flights approximately follow the linearized dynamics in Eq.~\eqref{equ:linearizedSystem}. This predicts a damped oscillation if unmodeled aerodynamic drag is taken into account. However, the amount of inherent aerodynamic damping is unlikely sufficient to rapidly stabilize the system.

\subsubsection{Closed-loop control with MOCAP feedback}
To evaluate the controller's performance independently from the estimation, the MOCAP was employed to compute $\phi_0$, $\phi_1$, $\dot{\phi}_0$ and $\dot{\phi}_1$ in real-time. The groundtruth feedback was wirelessly transmitted to the robot for real-time control. Three flights were performed. The state estimates were also present, but not employed for flight control.

\subsubsection{Closed-loop onboard estimation and control}
Three flights were performed with both the onboard estimation and onboard feedback for control simultaneously. In this case, the flight stability depends critically on the accuracy of the estimates.

\subsection{Leader-Follower Experiments}
Finally, to demonstrate that the proposed methods can be applied to realize cooperative transport of a suspended payload by two robots in a practical scenario where the payload is too heavy for a single robot. The fixed-point was substituted by another Parrot Mambo drone as shown in Fig.~\ref{fig:cooperHoverPhoto}. We implemented a standard PID controller to the leader to control its position. The setpoint attitude of the leader was adjusted according to the nominal equilibrium condition provided by Eq.~\eqref{equ:reduced_manipulator}. This brought some slight movement to the leader's position. Three flights were performed.

\begin{figure}[tbp]
    \centering
    \includegraphics[width=.48\textwidth]{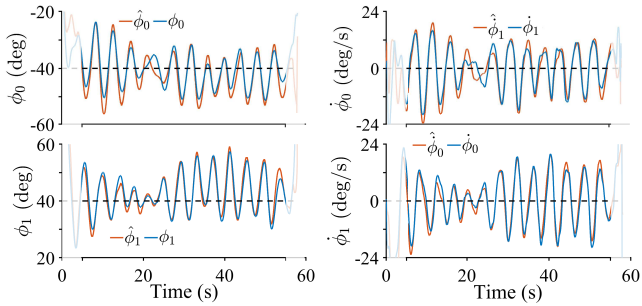}
    \caption{Plots showing $\phi_0$, $\phi_1$ and their estimates from open-loop flight.}
    \label{fig:noControl0822_1Video}
\end{figure}
\begin{figure}[tb]
    \centering
    \includegraphics[width=.48\textwidth]{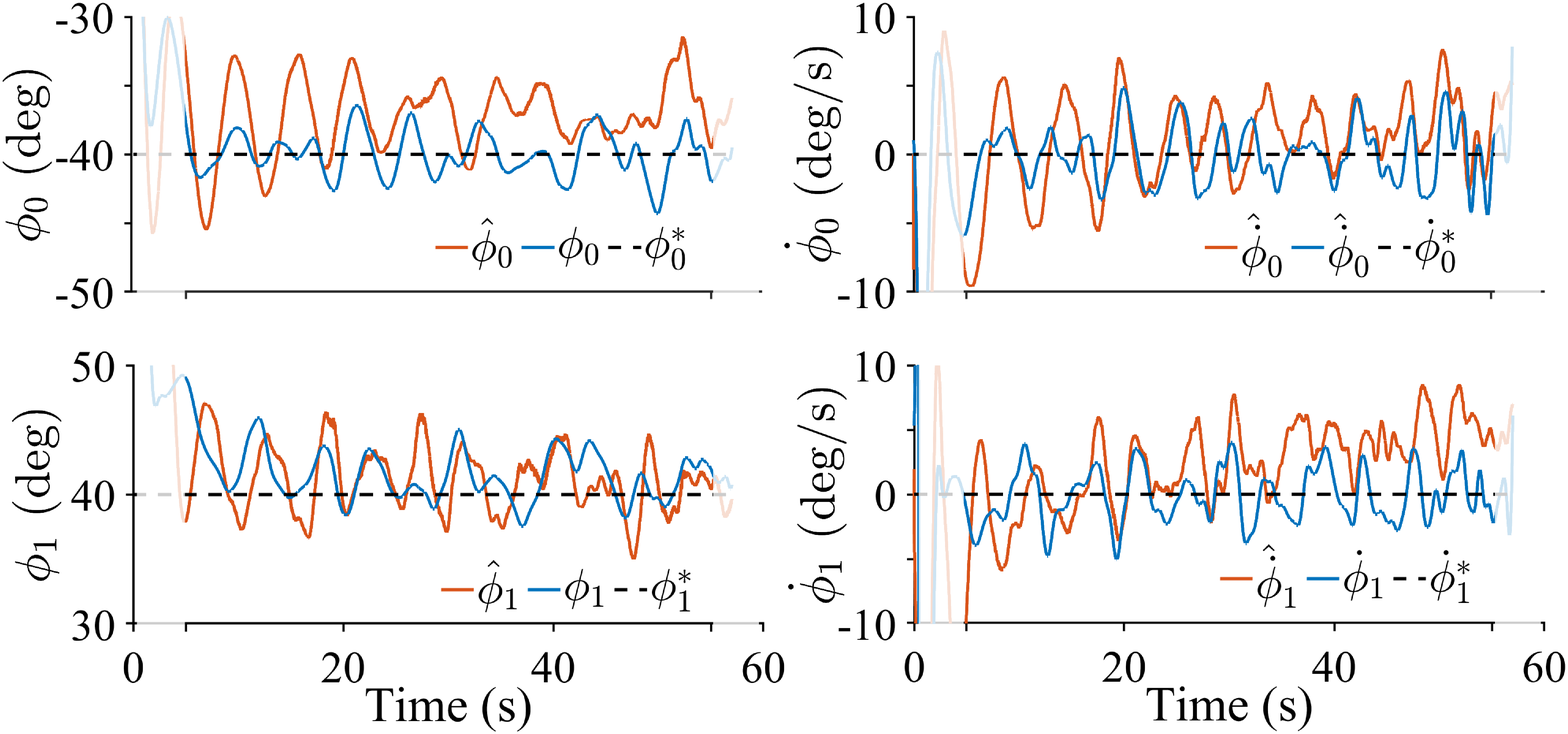}
    \caption{Plots showing $\phi_0$, $\phi_1$ and their estimates from controlled flight with MOCAP feedback.}
    \label{fig:udpControl0822_1Video}
\end{figure}

\subsection{Evaluation of Estimation and Control Methods\footnote{Example flight videos available as supplemental materials.}}

\subsubsection{Estimation results}
Fig~\ref{fig:noControl0822_1Video} shows the estimated angles and angular rates (denoted by $\hat{\cdot}$) against the groundtruth measurements from MOCAP from one sample of the open-loop flights. It can be seen that, without the corrective control terms, both angles oscillate around the setpoints with the amplitudes of approximately $~10-15^\circ$. The rates display a similar oscialltion of up to 20 deg/s. Despite the significant and rapid variation, the estimates track the groundtruth closely, advocating the performance of the estimator.
\begin{figure}[tb]
    \centering
    \includegraphics[width=.48\textwidth]{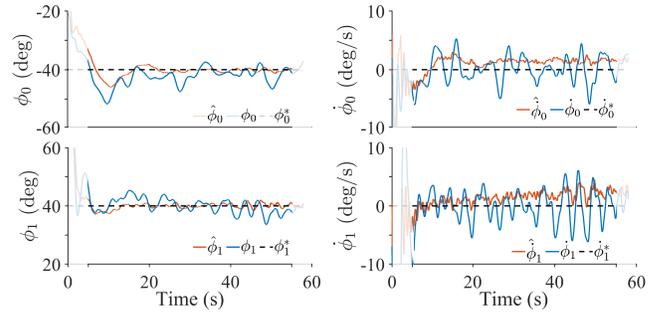}
    \caption{Plots showing $\phi_0$, $\phi_1$ and their estimates from controlled flight with onboard estimated feedback.}    \label{fig:FCspukfControl0822_4Video}
\end{figure}
\begin{figure}[tb]
    \centering
    \includegraphics[width=.48\textwidth]{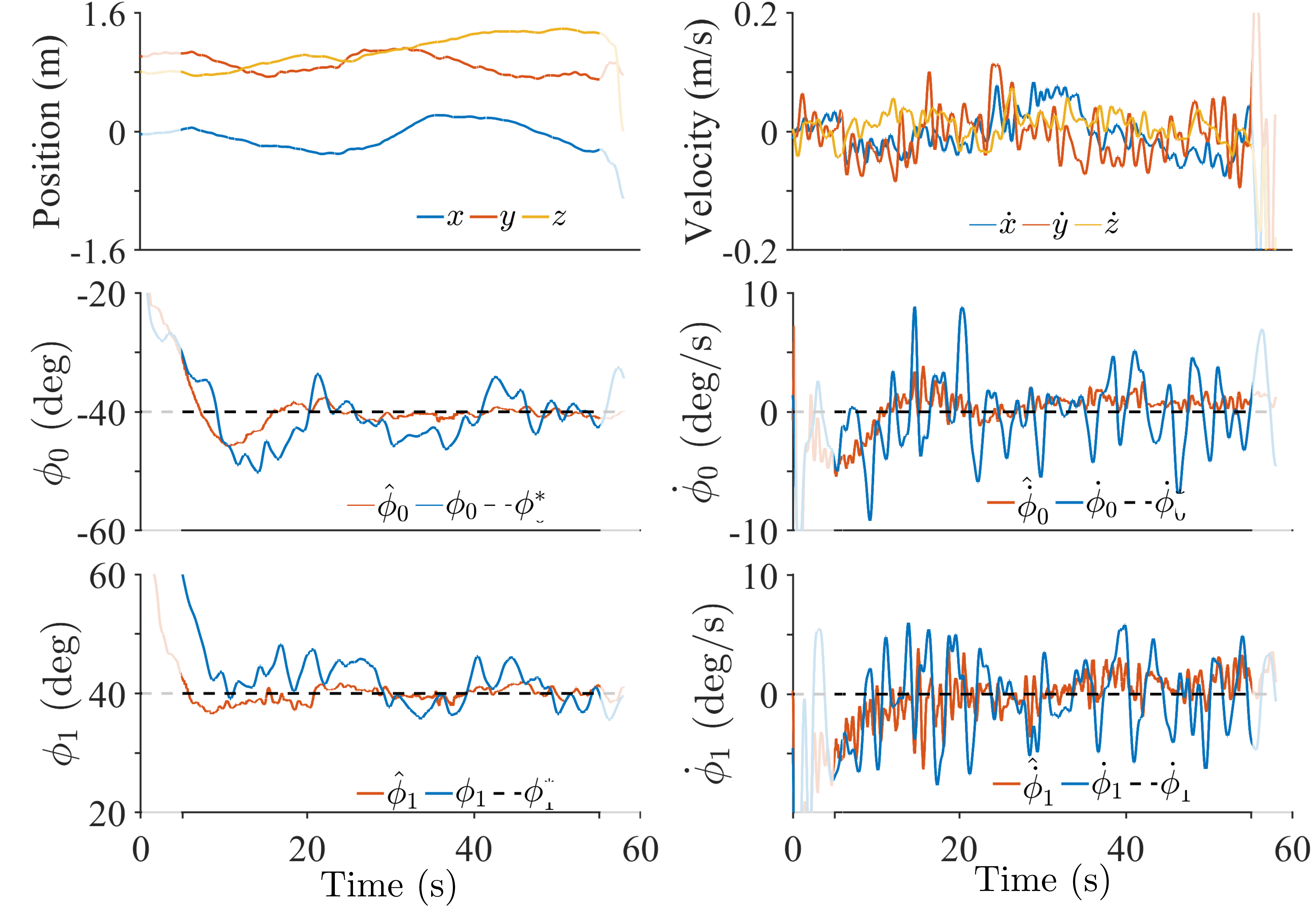}
    \caption{Plots showing (i) position and velocity of the leader robot; and (ii) $\phi_0$, $\phi_1$ and their estimates from controlled flight with onboard estimated feedback featuring two robots.}
    \label{fig:twoRobots0823_3Video}
\end{figure}
Fig.~\ref{fig:udpControl0822_1Video}-\ref{fig:twoRobots0823_3Video} illustrate example estimated results from the feedback controlled flights without and with the leader robot. Similar to Fig.~\ref{fig:noControl0822_1Video}, no significant estimation error is observed. However, some slight offsets of up to $5^\circ$ between $\hat{\phi}_{0,1}$ and $\phi_{0,1}$ can be seen in a few occasions. This marginal steady-state estimation error (as observed in Fig.~\ref{fig:udpControl0822_1Video}) is likely caused by uncertain model parameters (such as cable's length) or the inaccuracy of thrust model. A linear sensitivity analysis around the nominal state suggests that a only $5\%$ error in the thrust model would result in up to 15$^\circ$ and $20^\circ$ mispredictions of $\phi_0$ and $\phi_1$. With the significant change in battery levels during flight, the performance of the estimator is inevitably affected.
%the open-loop experiment, we see that $\phi_1$ and $\phi_0$ (blue lines) are kept oscillating within ranges of 30 deg and 20 deg, respectively; And their angular rates are varying between about -20 to 20 deg/s. All are in a sinusoid-like form, these observations highly support our theoretically conclusion in Subsection~\ref{subsec:LocalStability}, which is the system without control is locally marginally stable. This is the baseline of further analysis for both estimation and control.

\begin{figure}[tb]
    \centering
    \includegraphics[width=.48\textwidth]{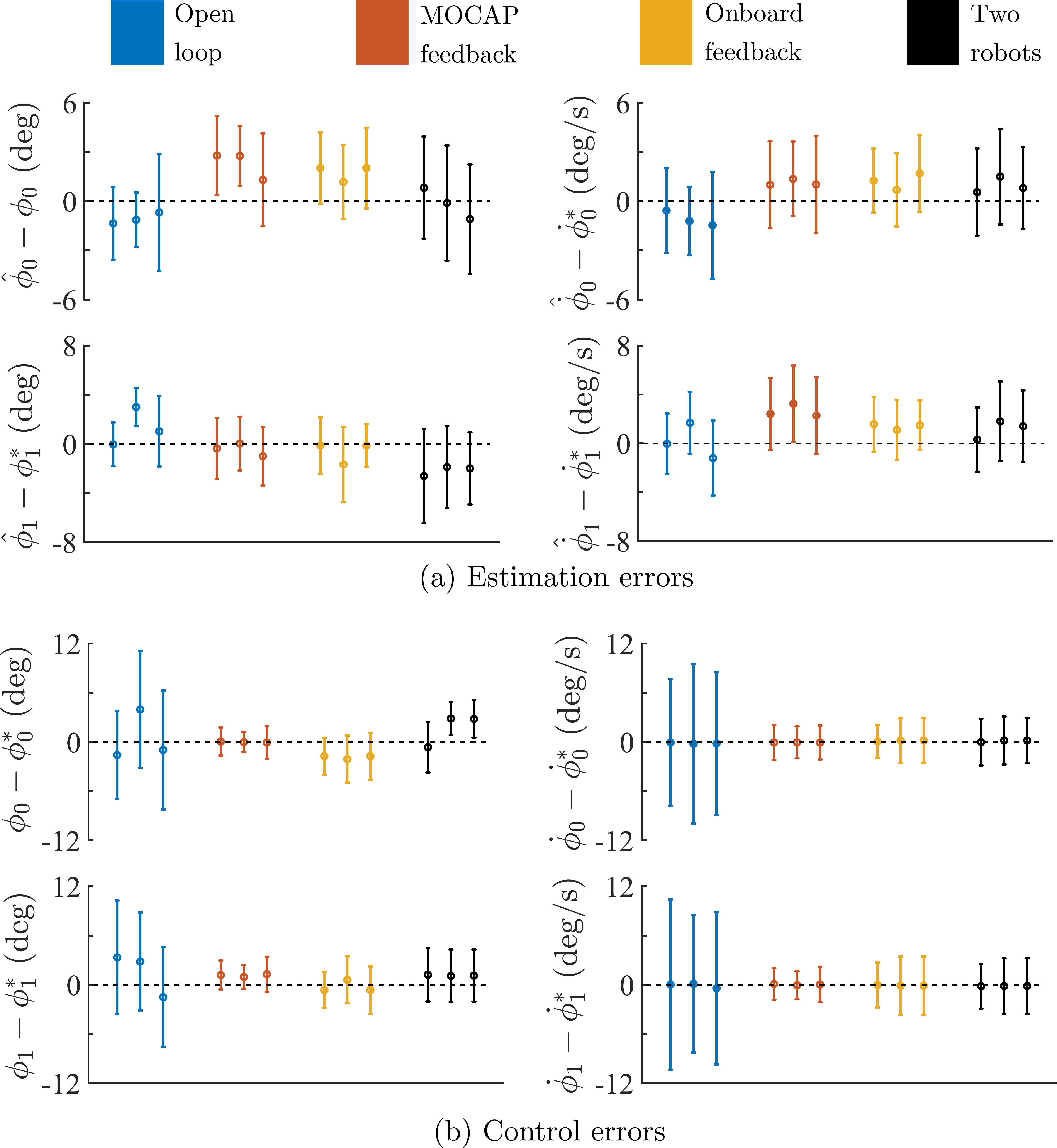}
    \caption{ (a) Means and standard deviations of the estimation errors calculated from $t=5-55$ s. (b) Means and standard deviations of the  control errors from $t=20-55$ s to exclude the transient behavior.}
    \label{fig:estimationRMSE}
\end{figure}

Quantitatively, Fig.~\ref{fig:estimationRMSE}(a) verifies that no significant difference between the estimation errors is visible through all 12 flights from four sets of experiments. The estimation errors are generally less than 5$^\circ$ or $3$ deg/s. Nevertheless, errors from flights with two robots feedback tend to be higher than others, perceivably due to the violation of the assumed quasi-static conditions.

%Again, look into Fig~\ref{fig:noControl0822_1Video}, the estimated states (orange lines) are very close to the ground truth; And their RMSEs are all less than 3 deg (or deg/s), as shown in Fig~\ref{fig:estimationRMSE}, where the RMSEs of EXP~3 and EXP~4 are also plotted, and each experiment was repeated 3 times. We see that statistically, the RMSEs are all less than 5 deg (or deg/s). Overall, the estimation algorithm are good enough for controller to stabilize the system.

%\begin{figure}[tbp]
 %   \centering
 %    \includegraphics[width=.48\textwidth]{figures_overleaf/estimationErrorBar_v3.pdf}
 %    \caption{Means and standard deviations of the state estimation. }
 %    \label{fig:estimationRMSE}
 %\end{figure}
 %\begin{figure}[tbp]
 %    \centering
 %    \includegraphics[width=.48\textwidth]{figures_overleaf/control20SecErrorBar_v3.pdf}
 %    \caption{Means and standard deviations of ground truths comparing with references. % The data are all trimmed from 20 to 55 seconds because we consider the steady-state error,   %and there are three trails for each experiment.}
  %   \label{fig:control20SecErrorbarEvaluation}
 %\end{figure}
\subsubsection{Flight control performance}

To examine the performance of the flight controller, first we consider one of the open-loop flights shown in Fig. \ref{fig:noControl0822_1Video}. Comparing groundtruths of $\phi_0$ and $\phi_1$ from these plots against those from closed-loop flights in Fig. \ref{fig:udpControl0822_1Video}-\ref{fig:twoRobots0823_3Video}, it is evident that the controller dramatically reduces the oscillation and angular errors from $>10^\circ$ to $<5^\circ$. Likewise, the plots of the angular rates show a substantial decrease in the angular rates from $\approx 20$ deg/s to $<5$ deg/s when the controller is employed. The outcomes in Fig.~\ref{fig:FCspukfControl0822_4Video} and \ref{fig:twoRobots0823_3Video} verify that small estimation errors are not detrimental to the system's stability. More comprehensive comparison between all experimental sets is provided in Fig.~\ref{fig:estimationRMSE}(b). As anticipated, the angular errors from flights with the MOCAP feedback are lowest---generally below a 2-3$^\circ$. This is because the MOCAP feedback is not susceptible to the inaccuracy of the estimates. Furthermore, a closer inspection into the dual-robot flights there is no significant difference compared to the single-robot cases even the leader breaks the near hovering assumption as shown in the position and velocity plot in Fig.~\ref{fig:twoRobots0823_3Video}. Therefore, it is reasonable to conclude that the proposed strategies can be effectively applied to the scenario with two robots with low-speed maneuvers, despite the use of the fixed-point assumption in the formulation.

%Regarding the controller, we refer to EXP~2, the experiment of control with feedback from a motion capture system, as plotted in Fig~\ref{fig:udpControl0822_1Video}. During 5 to 55 seconds, the mean values of $\phi_1$ and $\phi_0$ are 40.8 deg and -40.1 deg (note that our references are $\phi_1^{*} = 40$ deg, $\phi_0^{*} = -40$ deg), respectively. And mean values of $\dot{\phi}_1$ and $\dot{\phi}_0$ are 0.06 deg/s and 0.02 deg/s, respectively. Reminding the open-loop case we just mentioned, the effectiveness of controller is obvious. The cases of using feedback from the onboard estimator, EXP~3 and~4, the states' mean values and standard deviations are also decreased at least about a half when comparing with the open-loop experiments, as illustrated in Fig~\ref{fig:control20SecErrorbarEvaluation}.

%Last but not least, in the experiments of onboard estimation and control with two robots, even the leader is moving slowly, as depicted in the example shown in the first row of Fig~\ref{fig:twoRobots0823_3Video}, the system still shows comparable behavior to the experiments with using a fixed point as the leader as one can refer to Fig~\ref{fig:estimationRMSE} and~\ref{fig:control20SecErrorbarEvaluation}, this supports that the onboard combination of estimation and control works properly for formation stabilization by two aerial vehicles.

\section{Conclusion and Future Work} \label{sec:conclusion}
We presented an estimation and control strategy to address the formation control problem of two flying robots with a suspended payload. Treating the robots as a leader and a follower, the dynamics of the leader-payload-follower system is derived and the formation control problem is translated to a control of cables' angles. The proposed EKF-based estimator allows the relative position of the follower robot to be estimated using only onboard IMU measurements. With a simple feedforward PID controller, the proposed strategy has been extensively verified with flight experiments. The results demonstrate that the developed methods, which require minimal computational power, are capable of controlling the system by stabilizing the cables' angles such that they converge to the desired values with an accuracy of a few degrees. 

While the strategy has been shown feasible in practice, it is not without limitations. Thus far, the methods rely on two major simplifying assumptions, the neglect of the inertial effect of the leader and the planar consideration. Future work will take into account the acceleration of the leader to eliminate the quasi-static assumption and then extend the work to account for non-planar dynamics.
%%%%%%%%%%%%%%%%%%%%%%%%%%%%%%%%%%%%%%%%%%%%%%%%%%%%%%%%%%%%%%%%%%%%%%%%%%%%%%%%

\bibliographystyle{IEEEtran}
\bibliography{bibtex}
\end{document}